# Using Automated Vehicles Operational Data to Confirm Safety and Anticipate Threats

Riccardo Donà European Commission, Joint Research Centre (JRC), Ispra, Italy, riccardo.dona@ec.europa.eu

Espedito Rusciano, Dutch Vehicle Authority (RDW), Zoetermeer, Netherlands, erusciano@rdw.nl

Germana Trentadue, Piksel, Milan, Italy, germana.trentadue@ext.ec.europa.eu

Anastasios Tsakalidis, European Commission, Joint Research Centre (JRC), Ispra, Italy, anastasios.tsakalidis@ec.europa.eu

Maria Cristina Galassi, European Commission, Joint Research Centre (JRC), Ispra, Italy, maria-cristina.galassi@ec.europa.eu

## Abstract

*European Union (EU) policymakers adopted revolutionary data collection provisions for Automated Driving Systems (ADS) in the recently approved regulation that allows driverless vehicles to be operated on public roads. The framework is inspired by best practices developed at the United Nations Economic Commission for Europe (UNECE) level: the In-Service Monitoring and Reporting (ISMR); and by similar operational data collection regulatory approaches in nuclear energy production and transportation fields. The collection of real-world data will enable the competent safety authorities to gather the information needed to confirm the homologation safety target. Safety-relevant driving scenarios discovered during the real-world operation of a given ADS can also be stored in a scenario catalogue to investigate how other ADS types might have addressed such a traffic conflict. Moreover, lessons learnt deriving from the data collected can be shared among original equipment manufacturers (OEMs) and safety authorities. Ultimately, the ISMR is recognised as a necessary tool to properly tackle the challenges associated with ADS safety assessment given the number of unknowns that might remain undisclosed by leveraging the traditional homologation validation scheme only.*

## 1 Introduction

Driving automation is gradually transitioning from a technological dream into a reality since the recent introduction of regulations for the type-approval of vehicles featured with Automated Driving Systems (ADS). In fact, the UNECE regulation 157, firstly published in 2021 [1] and recently amended [2], establishes provisions for the approval of the Automated Lane Keeping System (ALKS): the first driving automation system moving beyond SAE J3016 Level 2 [3]. Albeit in an ALKS-featured vehicle a human driver is still required, the ALKS performs the driving task when engaged. Thus, the manufacturer has the legal responsibility once the automation system is activated (see **Figure 1**). Thus, the ALKS constitutes the first example of a *conditional automation* driving system, which contrasts with the *driving support features* of Level 2. One further step towards driverless vehicle automation was undertaken in Europe with the adoption of the EU Implementing Act 2022/1426 [4], which created the legal basis to type-approve a set of SAE Level 4 automated applications, namely: 1) hub-to-hub services, 2) robo-taxis, 3) urban shuttles and, 4) Automated Valet Parking (AVP).

**Figure 1.** SAE J3016 automation levels.

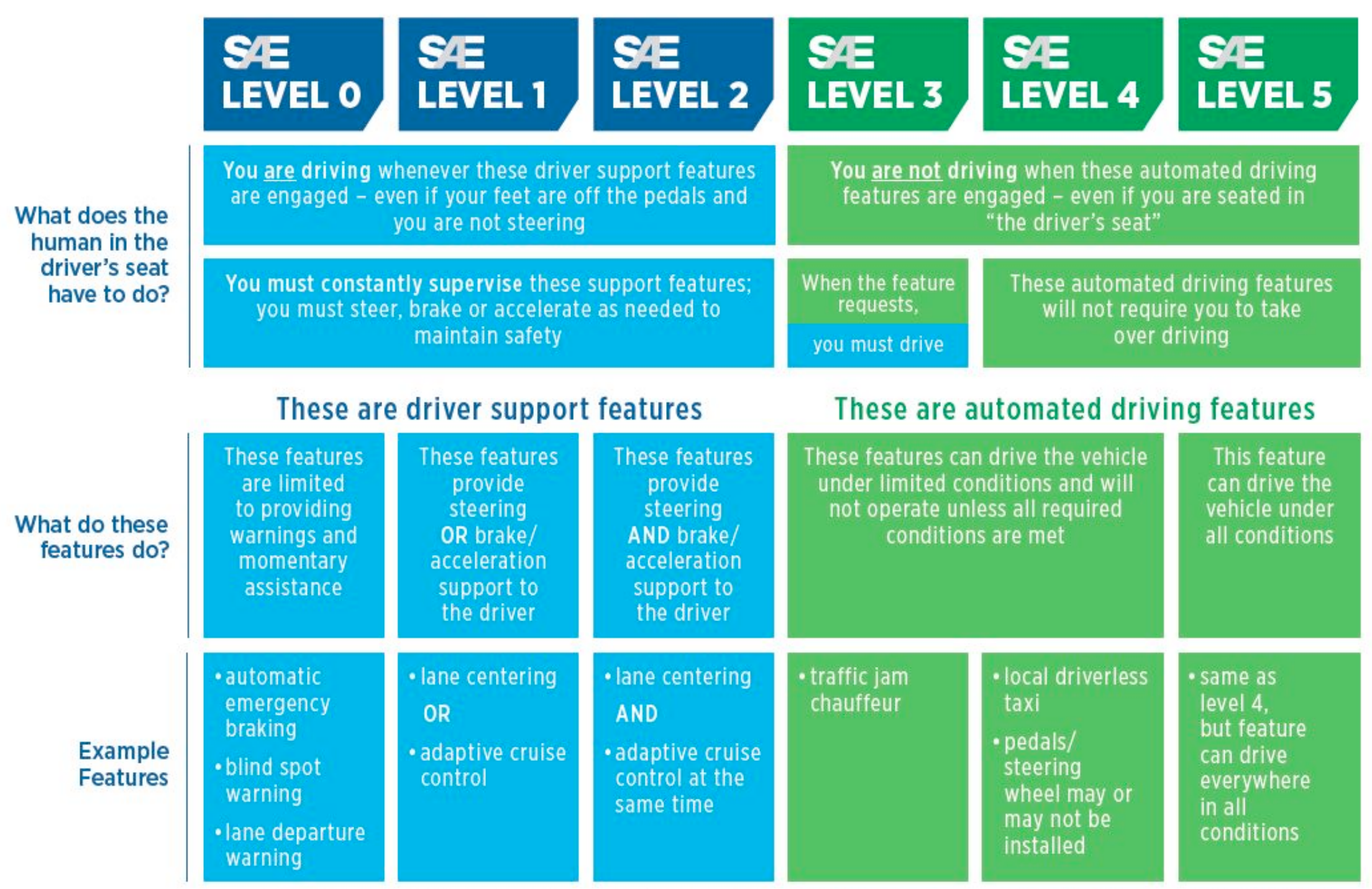

*Source:* Adapted from [3].

While drafting the technical requirements for the ADS homologation, authorities and policymakers were faced with the arduous task of designing approaches that would ultimately ensure the inherent safety of the automation technology and its interaction with the existing transportation network. It is indeed widely acknowledged that gathering such a confirmatory evidence via open road tests only would require billions of kilometres to be driven [5] thus making this validation method poorly scalable. Hence, the regulators' approach was to survey existing best practices and regulatory provisions within the other fields where safety-critical cyber-physical automation technologies have already been enforced. Those included: aviation, railway, and nuclear [6].

One key aspect separating the regulatory provisions within relevant fields analysed to the existing type-approval practices in the automotive industry was the collection of *operational data* during the in-use phase of a safety-critical technology after the certification of the system. In multiple reports [7,8], the learning phase made possible by the in-service data collection was deemed necessary to produce effective legislation that ultimately yielded to the safety level we witness nowadays. Such a finding is also particularly true for ADS given their technological complexity, which makes them very difficult to validate by leveraging the traditional type-approval scheme only.

In fact, ADS are subjected to varying in-service conditions since they do not operate in a controlled environment as other automation applications. Hence, operational unknowns are to be considered a normal aspect of each ADS design and lifecycle [9,10]. Additionally, the experience collected using one particular type of vehicle could be shared among other manufacturers to anticipate possible failures in case the need is recognised, thus introducing a proactive component into the safety assessment.

Based on the related experience analysed, the ADS policymakers devised a homologation scheme leveraging a *multi-pillar* approach that is schematically presented in **Figure 2**: the New Assessment/Test Method (NATM). In particular, the type-approval starts with an analysis of the Operational Design Domain (ODD), i.e., "the operating conditions under which a given ADS or feature thereof is specifically designed to function" [3]. Those include, for instance, roadway type, weather conditions, geographical regions, and time of the day. Based on the ODD, the requirements and the testing scenarios are derived which inform the audit, testing, in-service monitoring and reporting (ISMR), and the safety assessment phases.

**Figure 2.** New Assessment/Test Method.

*Source:* Own elaboration based on UNECE, VMAD [11].

The audit phase is concerned with an analysis of the manufacturer's Safety Management System (SMS), i.e., the collection of processes and methodologies that have been put in place to guarantee the ultimate safety of the product developed. The testing phase enforces a pool of testing environments which includes credible simulation toolchains [12], track tests, and real-world tests to combine the advantages of each tool. Eventually, the ISMR allows collecting the operational data once the vehicle has demonstrated to fulfil the requirements. More specifically, the ISMR has the main aim of:

- **confirming** the safety assessment *after* type-approval has taken place;
- **identifying** new scenarios potentially relevant for the ADS validation;
- **sharing** safety-relevant recommendations.

In the continuation of the paper, we give extensive details on the practical realisation of the ISMR. The focus is on the key role played by the data collection procedures in managing the unknowns that underline the ADS real-world operation. Such an operational experience provides confirmatory safety evidence that it would have been otherwise impossible to collect cover using traditional testing methods only.

The paper is organised as follows. In Section 2, we present a brief literature review of the approaches considered when establishing the EU and UNECE provisions. In Section 3, we introduce the ISMR framework more extensively. Eventually, in Section 4 we present the conclusion and discuss the way forward.

## 2 Literature review

Gathering operational evidence concerning the performance of a system during real-world service is an extremely valuable information. Such a data collection approach constitutes one of the cardinal principles of many SMS within several fields. This Section briefly summarises the practical implementation of monitoring and reporting mechanisms within the nuclear, aviation, and road transportation fields to the end of providing the necessary background to fully grasp the relevance of the NATM ISMR.

### 2.1 Nuclear Power Plant

Monitoring and reporting tools are key components of nuclear power plant safety [7,13]. The International Atomic Energy Agency (IAEA) provides nuclear plant operators with guidelines for the collection and distribution of plants' monitoring data [14]. The guidelines contain a list of practical examples, which emphasises the roles and responsibilities of the management within the plant installation. From the guidelines, all events outside normal operation shall be analysed and the corresponding lessons learnt shall be distributed to other operators and authorities. Concerning the reporting phase, safety-relevant occurrences shall be immediately notified to the responsible authority. After the notification phase, a thorough report is expected to provide further clarification on the nature of the event without assigning blame but focusing instead on the lesson learnt. Eventually, following a screening action, the relevant event will undergo a dedicated *investigation* phase.

Albeit nuclear reporting relies on a different set of data elements and events to be reported with respect to the Automated Vehicles (AV), the experience matured in the nuclear field substantially inspired the drafting of the NATM guidelines for what concerns the ISMR tool. In fact, the continuous monitoring phase and the reporting of relevant occurrences are also fundamental principles of the ISMR. Nonetheless, the nuclear best practices provide extensive details on the practical implementation of the mechanisms, due to the long-lasting experience, to a level of detail that is currently not yet achieved in the AV field.

## 2.2 Aviation

Various transportation sectors are supported by robust and well-established monitoring and reporting systems. In particular, aviation, railway, and maritime can leverage software applications derived from European Coordination Centre for Accident and Incident Reporting Systems (ECCAIRS) [15]. ECCAIRS provides the users with a predefined taxonomy and pre-compiled templates to facilitate the reporting. The database has been developed since 1974, initially for the aviation field only and extended to the railways and maritime sector afterwards.

Concerning aviation, the data collection procedures follow a scheme leveraging a monitoring action coupled with reporting requirements. Provisions and guidelines [16] to help operators and manufacturers to instantiate a data monitoring exercise are given by the European Aviation Safety Agency (EASA) by means of the Flight Data Monitoring (FDM) scheme.

Reporting practices are established as well through a dedicated set of regulations [17] and via the establishment of a database[1]. Reporting is foreseen for a list of occurrences and within different timescales depending on the safety relevance of the incident/accident. For instance, serious accidents which require immediate reaction shall be notified within 72h. Moreover, not only organizations (for instance fleet operators and manufacturers) but also avionic professionals (for instance, maintenance and ground operation personnel) have obligations to report.

Overall, based on the information collected, EASA can carry out statistical analysis and monitor the evolution of safety. This results in the annual issue of the "Annual Safety Review" where safety-related events are reported. The report contains extremely valuable information since the authority can have a complete picture of the safety level achieved by the aircraft operators. Moreover, based on the information collected, the agency can issue safety recommendations as in the 2022 Annual Safety Recommendation Review [8].

Eventually, the recent introduction of a standardised risk classification system, the European Risk Classification Scheme (ERCS) [18], supports the effectiveness of the data collection by providing a set of severity indexes for a list of critical events and associated probabilities.

## 2.3 Road Transport

Road transport related occurrences which involved injuries or fatalities shall be reported to the CARE [19] database in the EU by the Member States. The road transport reporting mechanism is inherently more limited than the aviation reporting tool since the occurrences where police investigation was not required (e.g., damage-only accidents and minor incidents) are not reported. Moreover, harmonisation issues historically affected the CARE database. In this framework, the use of the Common Accident Data Set (CADaS) [20] taxonomy is recommended since 2011 to foster harmonisation of data elements to be reported.

With the introduction of a standardised Event Data Recorder (EDR) and Data Storage System for Automated Driving (DSSAD) for vehicles equipped with ADS, we can expect a major benefit also for European road authorities not only for manufacturers and insurance companies [21] in getting objective evidence to analyse safety critical events. Nonetheless, EDR/DSSAD tools remain limited in the information they can provide due to the limited number of triggering conditions and their main purpose as devices for liability assignment thus not stimulating a just culture framework.

[1] https://aviationreporting.eu/

In particular, EDR/DSSAD will not afford the analysis of near misses [22,23] (i.e., events that did not result in damages or injuries but had the potential to do so) nor there will be the possibility of proactively anticipating critical events by using those tools only. Near misses are indeed indicated to be between several hundred and a few thousand times the number of actual critical scenarios as reported in many safety pyramids, an example of which is given in **Figure 3**. Currently, the exact ratio between near misses and critical occurrences remains unknown due to the unavailability of recorded data. Nonetheless, ADS-collected data might contribute to build additional awareness for effective, anticipative, corrective actions so that major events could be prevented and avoided.

**Figure 3.** Safety pyramid.

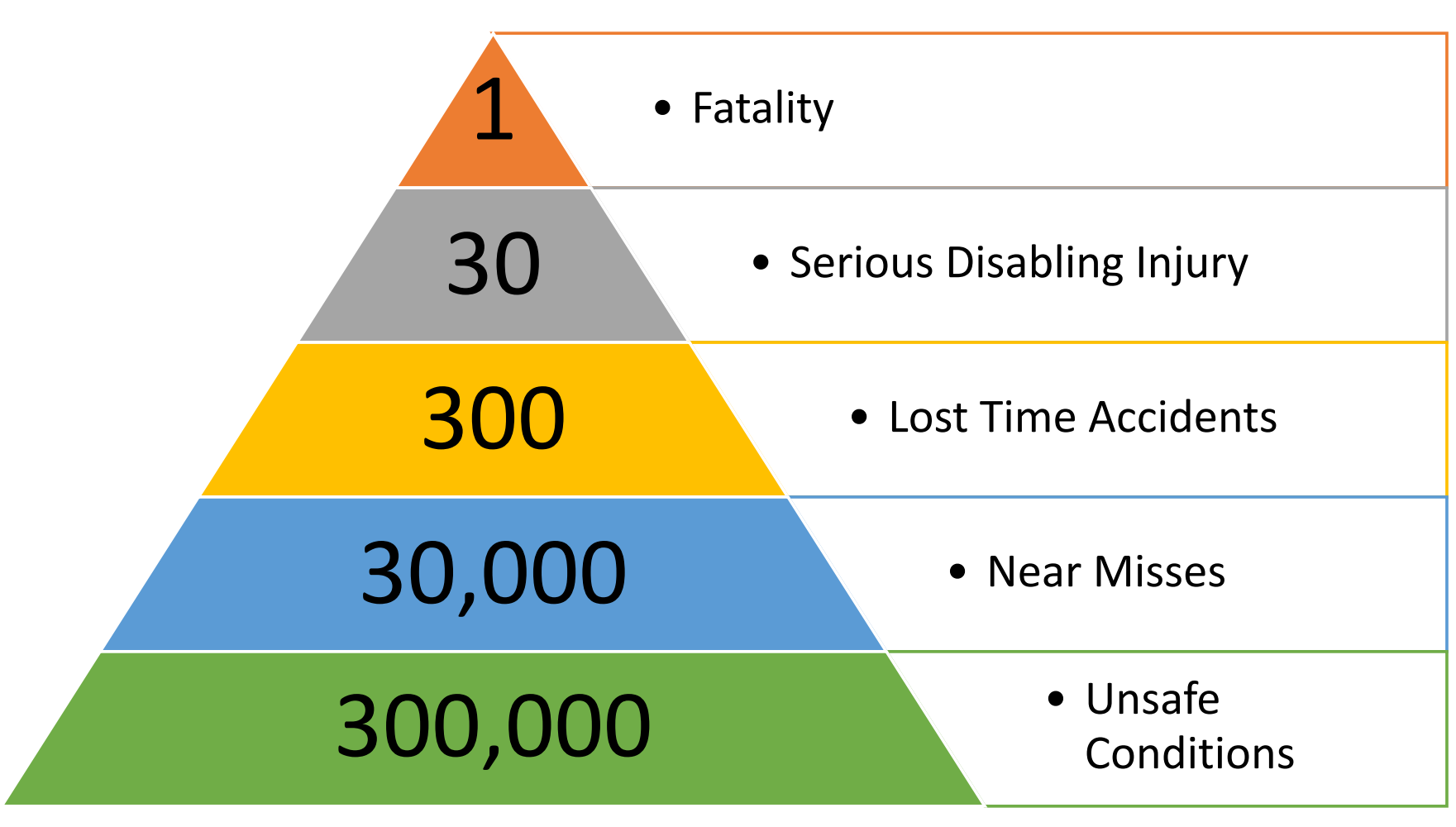

*Source:* own elaboration based on [24].

Concerning the specific use case of AVs, the collection of data mandated by the California Department of Motor Vehicles has already been demonstrated to provide useful information to California authorities as summarised by Favarò et al. in [25]. Similarly, the United States National Highway Traffic Safety Administration standing order for ADAS and ADS reporting [26] is starting to provide some preliminary information about the interaction between ADS and the existing transportation network.

## 3 Methodology

The NATM scheme presented in **Figure 2** highlights the key role played by the ODD in the ADS safety assessment. Based on the foreseen operational environment, the exposure to a set of critical scenarios can be estimated and the corresponding tests can be performed either virtually, for instance via simulation, or physically. Nonetheless, there are unknown unknowns that can only be tackled by the collection of operational data once an ADS has demonstrated to comply with all the requirements mandated by the NATM.

Based on the literature reviewed in Section 2 and based on the limited coverage that public road testing can deliver [5] due to the black-boxness underlying the artificial intelligence (AI) behind AVs, policymakers have realised the strong need for an ISMR tool to minimise the residual unknown. A key issue that promptly arose was the scalability of a data collection procedure. The ISMR cannot, in fact, be grounded on the collection of all the data generated by the complete ADS vehicle fleet. In fact, if ADS-equipped vehicles were to be required to store all the raw camera frames, assuming for example just a front-facing 1080p camera recording at 24 frames per second, that would generate a $1.19 \cdot 10^3\ Mbps$ video stream corresponding to about $537\ GB$ of data per hour of operation per vehicle. Naturally, more complex sensor setups will result in higher figures. An example chart to evaluate the video stream bitrate for different camera resolutions and frame rates is shown in **Figure 4**.

**Figure 4.** Raw camera video stream data generated.

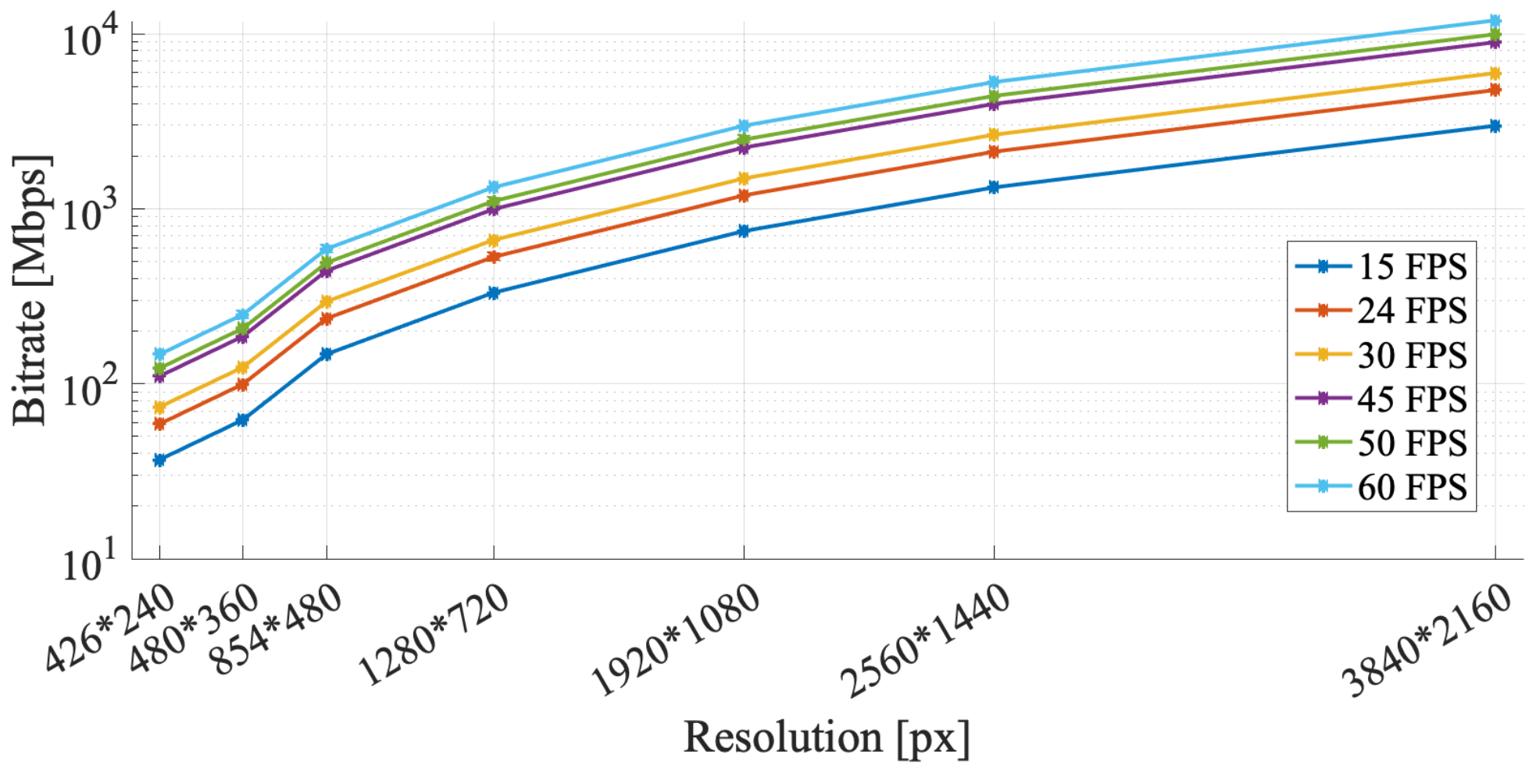

*Source:* Own computations.

The data collection approach should instead be capable of balancing the right degree between proactivity and severity coverage but, at the same time, be scalable enough for the road transport market. Eventually, ISMR should provide protection for confidential information released by the ADS manufacturer and anonymity for the reporting sources.

To cover the mentioned use cases, the conceived ISMR realisation for AVs leverages the three main methods:

- monitoring;
- reporting;
- investigation.

A schematical representation of the methods with respect to the degree of anticipation vs. the severity of the covered occurrences is graphically represented in **Figure 5**.

**Figure 5.** ISMR scheme.

*Source:* own elaboration based on [11].

The remaining of this Section describes the individual methods in detail.

## 3.1 Monitoring

The monitoring pillar is mainly concerned with the *continuous collection* of relevant data elements during normal ADS operation. The goal of the monitoring tool is to provide a *proactive* approach to collect confirmatory safety evidence of the SMS audit results according to the NATM framework **Figure 2**. The goal is achieved by leveraging a data recovery strategy coupled with a data retention strategy and a protection policy.

The monitoring tool will prove particularly useful for the:

- identification of **emerging trends** (e.g., degrading braking capabilities, more frequent near misses). Those trends might reveal, for example, vehicle-specific issues due for instance to the malfunctioning of a sensor that results in harsher braking manoeuvre with respect to the rest of the fleet;
- identification of **unusual or unsafe circumstances** which do not result in a trigger for the reporting. An example of that is the identification of operating conditions within the ODD that generate a higher risk level with respect to the remaining portion of the ODD and that might necessitate further refinement for the ADS to reduce the risk;
- identification of events where the **ADS managed to prevent incidents/accidents** to foster positive lessons sharing. Those occurrences are associated with a successful mission completion despite the risk during the operation reaching a substantial level, thus signalling that the ADS proved particularly effective.

Research activities are ongoing to identify the right amount of data elements and recording specifications to provide a sufficient amount of evidence without overloading the authority's capacity of processing the data. As noted in **Figure 4**, recording all the raw video information is not a viable approach and processed sensor information should be used instead for the monitoring.

Moreover, based on the data elements, widely established surrogate safety models can be enforced to compute the risk associated with the scenario [27] and compare the driving behaviour of the ADS with human driving historical data. For instance, the Time-to-Collision (TTC) [23],

$$TTC = \frac{distance}{v_{ego} - v_{leader}} \quad iff \quad v_{ego} > v_{leader}, \tag{1}$$

where $v_{ego}$ is the velocity of the ego vehicle, $v_{leader}$ the velocity of the leader, returns the residual time between a rear-end collision if the conditions remain unchanged, provide useful means to evaluate the criticality of rear-end scenarios, as of **Figure 6**.

**Figure 6.** TTC scheme.

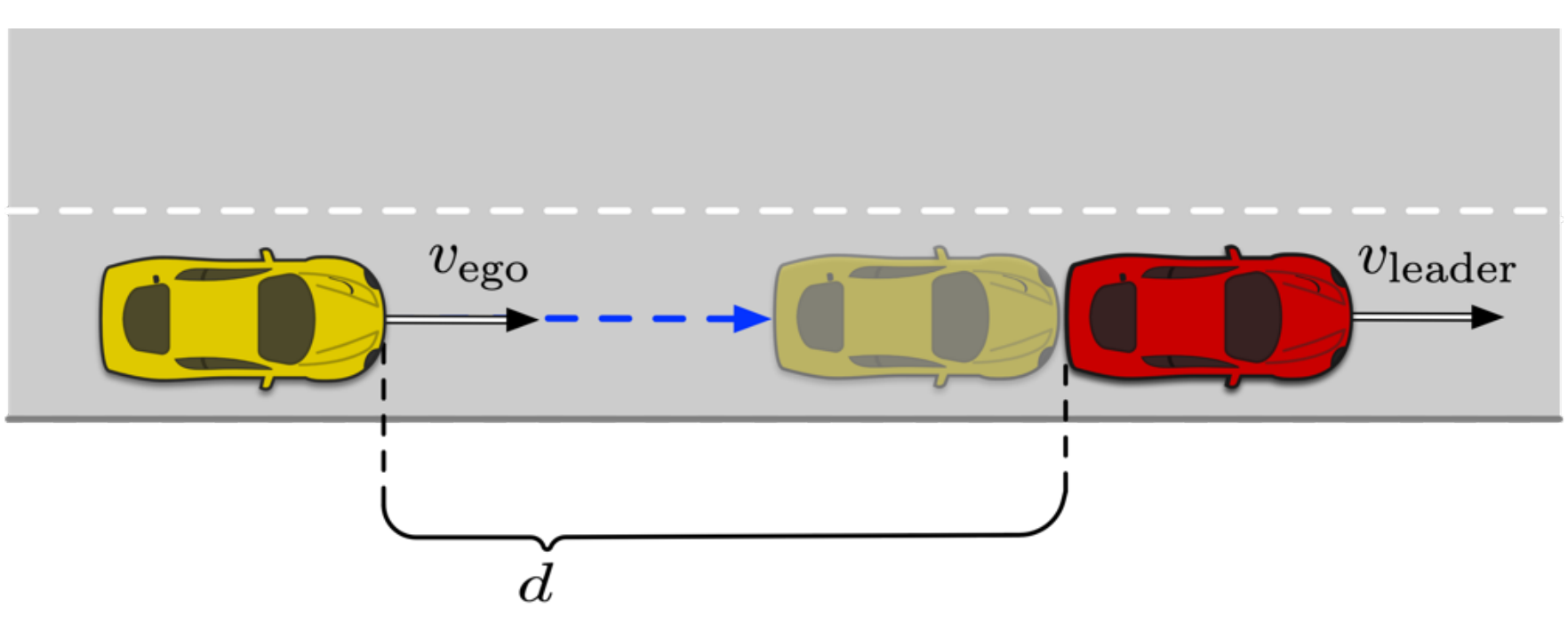

*Source:* own elaboration.

## 3.2 Reporting

The reporting pillar constitutes an *event-based* data collection methodology that is triggered by the happening of one particular occurrence among a set of events. Currently, the policymakers have listed a set of events which are reported in **Table 1**.

In particular, two approaches are foreseen by the current NATM framework:

- short-term reporting;
- periodic reporting.

Regardless of the periodicity, the two reporting schemes do not target assigning liability or attributing blame to the ADS manufacturer or operator. Instead, the reporting is envisioned as a tool aiming at stimulating a 'just culture' principle within the organisation.

**Table 1.** List of Reporting occurrences.

| Occurrence | Short-term [1 Month] | Periodic Reporting [1 Year] |
|---|---|---|
| **1. OCCURRENCES RELATED TO THE ADS PERFORMANCE OF THE DDT, SUCH AS** | | |
| 1.a. Safety critical occurrences known to the ADS manufacturer or OEM | X | X |
| 1.b. Occurrences related to ADS operation outside its ODD | X | X |
| 1.c. ADS failure to achieve a minimal risk condition when necessary | X | X |
| 1.d. Communication-related occurrences | | X |
| 1.e. Cybersecurity-related occurrences | | X |
| 1.f. Interaction with remote operator if applicable | | X |
| **2. OCCURRENCES RELATED TO ADS INTERACTION WITH FULLY AUTOMATED VEHICLE USERS, SUCH AS:** | | |
| 2.a. Driver unavailability (where applicable) and other user-related occurrences | | X |
| 2.b. Occurrences related to Transfer of Control failure | | X |
| 2.c. Prevention of takeover under unsafe conditions | | X |
| **3. OCCURRENCES RELATED TO ADS TECHNICAL CONDITIONS, INCLUDING MAINTENANCE AND REPAIR:** | | |
| 3.a. Occurrences related ADS failure | | X |
| 3.b. Maintenance and repair problems | | X |
| 3.c. Occurrences related to unauthorised modifications | | X |
| 3.d. Modifications made by the ADS manufacturer or OEM to address an identified and significant ADS safety issue | X | X |
| 4. Occurrences related to the identification of new safety-relevant scenarios | | X |

*Source:* EU ADS implementing act annex, [28].

Common to both approaches is the need to make the collected information available to all the contracting parties where the ADS vehicle type is operating but, at the same time, to retain industry-sensitive information confidentiality. Such a centralisation of the information collected is carried out by the instantiation of a centralised database. The database will retrieve the individual contracting parties' reported information and stores the individual entries after a screening exercise aimed at avoiding duplicates/finding relevant patterns in the occurrences as shown in **Figure 7**. Ultimately, the central authority will have an overarching understanding of the ADS functioning across all the contracting parties that will enable the same to issue safety performance reports such as [29].

**Figure 7.** Reporting mechanism.

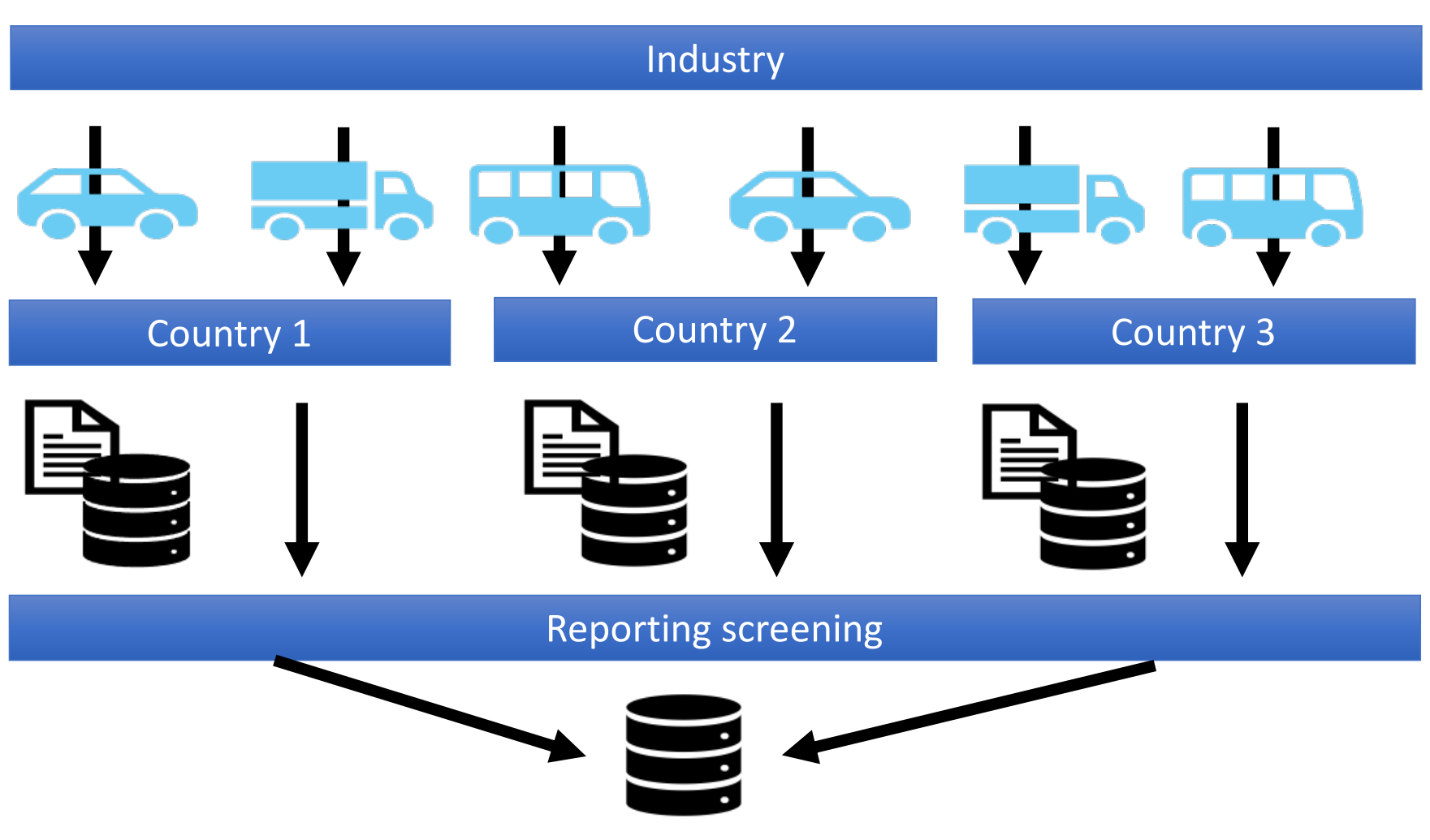

*Source:* own elaboration based on [11].

The remaining of this Section gives a detailed characterisation of the reporting provisions. It shall additionally be noticed, besides the short-term and periodic reporting, that voluntary reporting from any stakeholder involved in the ADS operation is encouraged.

### 3.2.1 Short-term reporting

The short-term reporting is a required action to be carried out by the ADS manufacturer/operator within one month following the occurrence of one of the events listed in the second column of **Table 1**. The aim of short-term reporting is to provide awareness concerning the risk that an ADS system might pose to the transportation network's safety, and that requires prompt remedial action. Albeit short-term reporting is a reactive tool since it follows the happening of a critical event, the lesson learnt might contribute to avoid analogous events in the future.

The short-term reporting scheme leverages two parts: a textual summary of the event and the ADS vehicle recorded data for later analysis. Currently, discussions are ongoing on the list of data elements to be provided[2] by means of a short-term reporting template. We claim that the introduction of a structured way of reporting the information, *i.e.*, by means of the aforementioned template, will increase the reporting significance by providing the authority with a consistence information layout. Such a provision will also prove more effective once a supporting taxonomy is devised such as is the case in aviation thanks to the ADREP taxonomy[3]. All the information contained in the report shall remain confidential to the safety authority receiving such material.

### 3.2.2 Periodic reporting

Periodic reporting aims at collecting regular information during ADS operation to provide evidence for the ADS safety and validate the ADS manufacturer safety claim. The report has to be provided on a yearly basis and contains aggregated data elements information concerning the whole fleet of a certain ADS type.

[2] https://wiki.unece.org/display/trans/VMAD-24th+SG3+session?preview=/186516577/188285355/VMAD-SG3-24-06%20VMAD-reporting-template_v1.1.docx

[3] https://www.icao.int/safety/airnavigation/aig/pages/adrep-taxonomies.aspx

Discussions are ongoing regarding the list of performance metrics to be provided by the ADS manufacturer or ADS operator to the competent safety authority[4]. Nonetheless, relevant metrics that should be reported are, for instance, the cumulative number of hours of operation, the cumulative driven mileage, and the number of disengagements. This way, the competent authority could have a reliable estimation of the occurrences' rate and anticipate hazards.

Another valuable piece of information that periodic reporting could deliver is related to *precursors* metrics. In fact, it is particularly important to recognise the statistical exposure of a system to certain risk levels. Such additional knowledge will enable the competent authority to establish a stronger link between the frequency of risky situations and critical events and thus be more anticipative. For example, in a dual manner to the TTC risk-based surrogate metric, the Time-Exposed-TTC (TET) [30] returns the cumulative time spent below a given TTC level. By being informed on the aggregated TET through the periodic reporting and the frequency collision events (TTC = 0) via the short-term reporting, the authority could devise the safety pyramid, as in **Figure 3**, for each ADS type thus mapping the exposure to risk (TET) into accident/incident frequency.

### 3.3 Investigation

The last pillar making up the ISMR framework is the investigation. The investigation is concerned with the timely response to a safety critical event by a competent body. For the investigation to take place, it is recommended that each contracting party nominates a dedicated authority to handle occurrences needing dedicated attention. The investigation authority shall, additionally, be independent of any interested party to ensure the impartiality of the analysis.

The investigation pillar will have the main task to unveil the reasons behind a serious accident. Following an investigation action, it is expected that remedial action is undertaken by the ADS manufacturer and corresponding safety recommendations are issued by the authority. The recommendations shall be made publicly available but, at the same time, no confidential detail should be disseminated.

## 4 Conclusions and Future Perspective

The paper has described the latest developments within the field of ISMR for AVs. The analysis focused on the synergies between the approach pursued in EU and UNECE regulatory and guideline approaches with respect to other related fields where ISMR tools are an established reality. Nonetheless, specific needs for the AV use case have been presented. Namely, the concerns related to getting the necessary statistical evidence to confirm the achieved safety level and the issues related to the scalability of an ISMR scheme.

The monitoring and reporting scheme devised for the AVs, albeit strongly inspired by existing practices in aviation and in the nuclear fields, needs in principle to balance a larger amount of systems to be monitored with respect to the other use cases analysed. Thus, the ISMR leverages three main methods: monitoring, reporting, and investigation. The contribution has also discussed the roles and responsibilities of the relevant authorities and contracting parties within the conceived ISMR framework.

Overall, the ISMR will allow to close the safety assessment loop with real-world operational data. We claim that such an approach will massively contribute to the identification of critical occurrences and the issuing of safety recommendations thus introducing a degree of proactivity into the AVs safety assessment. The process is part of a general transition towards an open regulatory framework that allows for more flexibility in adapting to upcoming technological development. Moreover, based on the operational feedback collected, a strong collaboration between manufacturers and authorities is envisaged.

Currently, the work is still continuing on the development of guidelines in an attempt to facilitate the instantiation of effective ISMR methodologies. In particular, detailed specifications for the data elements to collect and corresponding frequency prove to be challenging points to address. Additional issues include the definition of a taxonomy supporting the classification of the events and the different use cases between the fields of AVs which might necessitate dedicated ISMR approaches.

[4] https://wiki.unece.org/display/trans/VMAD-26th+SG3+session?preview=/192840175/192840653/VMAD-SG3-26-02%20periodic-reporting-template_v1.0.docx

## Acknowledgements

The authors would like to acknowledge the EU institutional funding for the possibility to establish the presented research activities and for the resources made available.

## List of abbreviations and definitions

| | |
|---|---|
| ADREP | Accident/Incident Data Reporting |
| ADS | Automated Driving System |
| ALKS | Automated Lane Keeping System |
| AI | Artificial Intelligence |
| AV | Automated Vehicle |
| AVP | Automated Valet Parking |
| CADaS | Common Accident Data Set |
| DSSAD | Data Storage System for Automated Driving |
| EASA | European Aviation Safety Agency |
| ECCAIRS | European Coordination Centre for Accident and Incident Reporting Systems |
| EDR | Event Data Recorder |
| EU | European Union |
| ERCS | European Risk Classification Scheme |
| FDM | Flight Data Monitoring |
| IAEA | International Atomic Energy Agency |
| ISMR | In-Service Monitoring and Reporting |
| JRC | Joint Research Centre |
| NATM | New Assessment/Test Method |
| ODD | Operational Design Domain |
| OEM | Original Equipment Manufacturers |
| SMS | Safety Management System |
| TET | Time Exposed TTC |
| TTC | Time To Collision |
| UNECE | United Nations Economic Commission for Europe |